\documentclass[11pt]{article}

\usepackage{amsmath,amsthm,verbatim,amssymb,amsfonts,amscd, graphicx, enumitem, times}
\usepackage{graphics}
\usepackage{centernot}
\usepackage{authblk}
\theoremstyle{plain}
\newtheorem{theorem}{Theorem}

\newtheorem*{remark}{Remark}

\newtheorem{definition}{Definition}

\usepackage[colorlinks,linkcolor=blue,citecolor=blue]{hyperref}

\usepackage[bottom]{footmisc}
\usepackage{caption}
\usepackage{subcaption}
\usepackage{helvet}  
\usepackage{courier}  
\usepackage{url}  
\usepackage{graphicx}  
\usepackage{multirow}
\usepackage{amsthm}
\usepackage{color}
\usepackage{MnSymbol}
\usepackage{makecell}
\usepackage{arydshln}
\usepackage{amsmath}
\usepackage[dvipsnames]{xcolor}
\usepackage{caption} 
\usepackage{natbib}
\usepackage{bbm}

\usepackage{textcomp}
\usepackage{wrapfig}
\usepackage{algorithm}
\usepackage{algorithmic}

\usepackage{csquotes}

\newcommand{\E}{\mathbb{E}}

\newcommand{\Tr}{{\rm Tr}}

\usepackage{minitoc}
\begin{document}

\title{Proof of a perfect platonic representation hypothesis}
\author{Liu Ziyin$^{1,2}$, Isaac Chuang$^{1}$\\
$^1$\textit{Massachusetts Institute of Technology}\\
$^2$\textit{NTT Research}}
\maketitle

\begin{abstract}
    In this note, we elaborate on and explain in detail the proof given by Ziyin et al. (2025) of the ``perfect" Platonic Representation Hypothesis (PRH) for the embedded deep linear network model (EDLN). We show that if trained with the stochastic gradient descent (SGD), two EDLNs with different widths and depths and trained on different data will become Perfectly Platonic, meaning that every possible pair of layers will learn the same representation up to a rotation. Because most of the global minima of the loss function are not Platonic, that SGD only finds the perfectly Platonic solution is rather extraordinary. The proof also suggests at least six ways the PRH can be broken. We also show that in the EDLN model, the emergence of the Platonic representations is due to the same reason as the emergence of progressive sharpening. This implies that these two seemingly unrelated phenomena in deep learning can, surprisingly, have a common cause. Overall, the theory and proof highlight the importance of understanding emergent ``entropic forces" due to the irreversibility of SGD training and their role in representation learning. The goal of this note is to be instructive while avoiding jargon and lengthy technical details.
\end{abstract}

\tableofcontents

\section{Introduction}

A recent line of research in AI has uncovered the emergence of universally aligned and structurally similar representations across different models -- a phenomenon referred to as the Platonic Representation Hypothesis \cite{huh2024platonic}.\footnote{We will use the adjectives ``universal" and ``Platonic" interchangeably. For example, a Platonic solution is also said to be a universal solution.} This hypothesis states that large models trained on larger and larger datasets will learn representations that are very close (if not identical) to each other, even if these models have different architectures and may even see different (but often related or paired) data.\footnote{In fact, Ref.~\cite{tjandrasuwita2025understanding} showed that if the two models are trained on completely unrelated data, there is no alignment.} For example, a language model taking in captions of an image has been found to have a similar representation to the representation of this image in a vision model. Interestingly and perhaps importantly, this similarity of learned representations occurs not only between different AI models but also between AI and biological brains \cite{yamins2014performance}, which suggests that understanding this phenomenon may also offer clues for understanding the biological brains and advancing neuroscience.

So far, the alignment between different models is rather weak empirically, and it is unclear whether the PRH is coincidental or reflects something truly profound. This raises three fundamental questions: (1) Is the phenomenology of mutual alignment a reflection of a truly fundamental and universal phenomenon? (2) If so, can we find an idealized scenario where pure and perfect alignment can be observed? (3) Can we have a theory that establishes this fact? We offer positive answers to all three questions. We will identify and solve a nontrivial mathematical model that can exhibit what we call a perfect Platonic Representation Hypothesis. Within this model, the precise cause of this phenomenon can be exactly identified and is due to none of the previously conjectured mechanisms -- and the actual cause is a universal mechanism of nature and physics: entropy maximization drives out-of-equilibrium systems towards universal states.

The model we will solve is called an embedded deep linear network (EDLN), which can be thought of as a deep linear network that is embedded in a larger network. We will find its exact solution (the exact form of global minimum) for what we call an ``entropic loss\footnote{Not to be confused with the ``cross-entropy loss."}" due to its very close analogy to nonequilibrium physics and thermodynamics. The entropic loss is a first-order approximation of the discrete-time nature and stochasticity of the SGD learning dynamics when modeling it using a continuous-time dynamics \cite{barrett2020implicit, smith2021origin}.\footnote{This loss is also known as the ``modified" loss. As we will discuss below, from a numerical-analysis perspective, the effect is related to the crude numerical integrators (such as SGD) not respecting the symmetries of the dynamics.} This entropic loss captures the implicit regularization effects of SGD training, and using this loss allows us to focus only on the geometrical properties of the loss function and avoid the difficult problem of actually solving the SGD dynamics. We will see that even with these simplifications, the solution is still very difficult to reach. In summary, the conclusion of this theory is that in this particular model we solve, the PRH is entirely due to the following mechanism, which we state in three equivalent terms for the ease of understanding for three different communities: 
\begin{itemize}[noitemsep,topsep=0pt, parsep=0pt,partopsep=0pt, leftmargin=25pt]
    \item implicit regularization (deep learning theory);\footnote{Because the result is due to interesting properties of the learning dynamics and cannot be understood through looking at the training objective alone. However, the mechanism relevant to PRH is the discretization and gradient-noise effect, not the continuous-time limit of SGD, which is the more common scenario in the study of implicit regularizations.}
    \item the growth of backward error of the numerical integrator (numerical analysis);
    \item the microscopic irreversibility of the learning dynamics (physics).\footnote{For this reason, we will use the words ``entropic" interchangeably with the ``implicit-regularization."}
\end{itemize}
Other interesting conclusions and broader implications are discussed in Section~\ref{sec: discussion}.

\section{Problem Setting}

\paragraph{Notation.} For a vector $v$, $v^\top$ denotes its transpose, $v^\top v$ its inner product, and $vv^\top$ its outer product. $\theta$ always denotes the entirety of trainable parameters, and $\nabla_v$ denotes the gradient with respect to $v$. When used without a subscript, $\nabla := \nabla_\theta$ is a shorthand for the gradient for all parameters $\theta$. Vectors without transposes are all viewed as column vectors; for example, $\nabla_v f(v)$ is a column vector and $\nabla_v^\top f(v)$ is a row vector. For a matrix $W\in \mathbb{R}^{d_1 \times d_2}$, the gradient of a scalar function $f$ with respect to $W$ is a matrix of the same size: $\nabla_W^\top f(W) \in \mathbb{R}^{d_1 \times d_2}$. Other notations are introduced in the context.

\paragraph{Platonic Representation Hypothesis.} We first define the idealized form of PRH. Let $x$ denote the input data. Consider two potentially different neural networks $A$ and $B$. Let $h_A=h_A(x)$ denote a hidden layer of network $A$ and $h_B=h_B(x)$ that of network $B$. It is possible for the input to the two models to take different forms; it is thus a good idea to consider two paired data points $(x_i^A, x_i^B)$, where $x_i^A$ is the data point seen by $A$ and $x_i^B$ is the data point seen by $B$. One example of such a dataset is a multimodal dataset with image and caption pairs, and net $A$ can be a CNN taking images $x_i^A$, and $B$ can be a language model taking language input $x_i^B$. For the purpose of modeling, our theory will consider the case where $x_i^A = Z x_i^B$ for some invertible matrix $Z$.

\begin{definition}
    Two layers $h_A$ and $h_B$ are said to be (perfectly) \textbf{aligned} if there exists a constant $c_0$ such that for any $x_1^A,\ x_1^B,\ x_2^A,\ x_2^B$, 
    \begin{equation}\label{eq: universal alignment}
        c_0 h_A(x_1^A)^\top h_A(x_2^A) = h_B(x_1^B)^\top h_B(x_2^B).
    \end{equation}
\end{definition}
The meaning of this definition is that when satisfied, the two models really encode the distance relationships between two different data points in the same way.\footnote{If we think of human knowledge as the relationship between different objects, then a perfect alignment implies that the two models really have the same ``knowledge" of the world.} This is an idealization of the PRH.\footnote{This definition of alignment corresponds to the CKA metric used in Ref.~\cite{huh2024platonic}.} Any distance between the two sides can be called the ``degree of disalignment." When two networks reach a perfect alignment between every possible pair of hidden layers, we say that these two networks satisfy the ``Perfect" PRH. A model parameter $\theta$ that satisfies this property is called a ``Platonic" or ``universal" solution.

A point worth raising is that this usage of the word and definition seems to imply that ``being Platonic" is a property of a single model, rather than a property of a pair of models. This criticism is valid because it is possible to imagine four models, A, B, C, and D, such that (1) A and B learn aligned representations, (2) C and D learn aligned representations, yet (3) A/B and C/D learned nonaligned solutions. This can indeed be a problem in general. However, as our proof will show, for the specific model we consider, running SGD makes all networks mutually perfectly aligned. Therefore, the \textbf{pairwise} alignment property can also be regarded as an \textbf{individual} property. It thus makes sense to say that a solution for A is ``Platonic" without referring to the reference network B. For now, we will simply assume that this is the case and say words like ``Platonic" solutions.

\paragraph{Model.} The model we study is what we call an \textbf{embedded} deep linear network, whose output for input $x$ is given by
\begin{equation}
    f_M(x) = M^OW_D\cdots W_1M^I x,
\end{equation}
where $W_i$ is the trainable weight matrix of the $i$-th layer, and $M^O$ and $M^I$ are arbitrary invertible matrices and are frozen during training. The $M$ matrices roughly model the layers coming before or after the deep linear network (thus the name ``embedded"). The \textbf{width} of this model is defined as the smallest row dimension of all the weight matrices. We will require the width to be larger than or equal to the rank of the target mapping.

We will denote the latent representation of layer $i$ with a superscript:
\begin{equation}
    h^i =  W_i ... W_1 M^I x.
\end{equation}
Because we will consider training two different networks, $M^O$, and $M^I$ will be different for the two networks and be written with a subscript $A$ or $B$. 

That this model reaches a Platonic solution is by no means trivial. We will explain in Section~\ref{sec: most solutions are not platonic} that almost all of the global minimizers of the loss function (to be defined in the next paragraph) are non-Platonic. That SGD only finds the Platonic ones is rather extraordinary. Also, while this model is a linear function of the input, its learning dynamics and loss landscape are regarded as good models of those of a nonlinear network. Thus, while it may not be a good model to capture the generalization capability or expressivity of neural networks, it captures the learning dynamics of neural networks rather well, as many prior works on deep linear networks have suggested \cite{saxe2013exact}. 

\paragraph{Data and loss function.} Let $\theta = (W_D,...,W_1)$ denote the set of all trainable parameters. The per-sample loss $\ell$ of a $D-$layer EDLN is: 
\begin{equation}
\ell(\theta, x)=||M^OW_D\cdots W_1M^I x-y||^2,
\label{eq:loss_deep_linear}
\end{equation}
where $y = V^*x + \epsilon$ for an i.i.d. zero-mean label noise $\epsilon$ with a full-rank covariance matrix $\Sigma_\epsilon$.\footnote{One might wonder if perfect PRH can be reached for a more complicated data distribution where $y = y(x)$ is a generic function of $x$. It turns out that PRH can no longer be achieved for generic labels. A key step in this proof requires the noise (or, prediction residual $\hat{y}-y$) to be independent of $x$. If the prediction residual is $x$-dependent, the global minimum will not be Platonic -- this is yet another way to break the PRH. Also, we stress that the labels having a nonvanishing and full-rank noise is quite crucial for the proof of the PRH. This is consistent with a key insight of this theory: it is noise in the gradient that determines representation learning.} Similarly, we denote the second moment of $x$ as 
\begin{equation}
    \Sigma_x  = \E[xx^\top] \,.
\end{equation}
$\ell$ is called a ``per-sample" loss because it is the loss for a single data point and the empirical loss we would like to minimize is its average over the training set: $L(\theta) = \E[\ell(\theta,x)]$. In the following, we consistently use $\E$ to denote the average over the training set (or over the distribution of data points if the training is online).

Now, we will train two networks (with potentially different depths and widths) on different ``views" of the same data, which can be seen as a mathematical abstraction of the multimodal experimental setting of Ref.~\cite{huh2024platonic}. The network $A$ is trained on $\mathcal{D}_{Z_A}$:
\begin{equation}
    \mathcal{D}_{Z_A} = \{(Z_A x_i, y_i)\}_i
\end{equation}
and network $B$ is trained on $\mathcal{D}_{Z_B}$,
\begin{equation}
    \mathcal{D}_{Z_B} = \{(Z_B x_i, y_i)\}_i,
\end{equation}
where $Z_A$ and $Z_B$ are arbitrary invertible matrices. Using the notation of Eq.~\eqref{eq: universal alignment}, this means that 
\begin{equation}
    x_i^A  = Z_A x_i,
\end{equation}
\begin{equation}
    x_i^B  = Z_B x_i.
\end{equation}
Note that with the three matrices $Z_A, M^O, M^I$, the global minimum of $L$ satisfies:
\begin{equation}
    W_D...W_1 = (M^O)^{-1} V^* (M^I)^{-1} Z_A^{-1},
\end{equation}
and at the global minimum $\theta^*$, the loss function value is determined by the noise $\epsilon$:
\begin{equation}
    L(\theta^*) = \Tr[\Sigma_\epsilon].
\end{equation}

\paragraph{Entropic loss due to SGD training.} Recent works showed that training with SGD at a finite learning rate $\eta$ and batch size is equivalent to minimizing the following objective \cite{barrett2020implicit, smith2021origin, ziyin2025neural}:
\begin{equation}\label{eq: entropic loss}
     \E[\ell(\theta, x)] + \eta {\E \|\nabla\ell\|^2},
\end{equation}
which states that SGD training implicitly introduces an additional regularization term equal to the expected norm of the gradient. Because this can be seen as an analogue of the entropic force\footnote{An entropic force is a force in physics that cannot be explained by looking at the energy function alone.}, we will refer to $\E[\|\nabla_\theta\ell\|^2]:= S$ as the ``entropy." We will consider the minimizers of the $\eta \to 0^+$ limit of the loss function, which is equivalent to finding the solution of 
\begin{equation}
    \min_{\theta: L(\theta) =\Tr[\Sigma_\epsilon]} {\E \|\nabla\ell\|^2},
\end{equation}
meaning that we are finding the minimal gradient norm solution $\theta$ with the constraint that $\theta$ is a global minimum of $L(\theta)$.

Note a key property of the entropy term:
\begin{equation}
    S = \sum_i \E\|\nabla_{W_i} \ell \|^2 =  \sum_i \E [ {\|\nabla_{h^{i+1}} \ell \|^2} {\|h^i\|^2} ].
\end{equation}
where ${\|\nabla_{h^i} \ell \|^2}$ is the representation gradient, a \textit{global} quantity, $\|h^i\|^2$ is the representation, a \textit{local} quantity. The entropy thus introduces a \textbf{coupling between the global gradient and the local representation}, encouraging both terms to be as simple as possible.\footnote{In the sense of having a small norm.} However, these two terms cannot be simultaneously minimized, and the tradeoff effects between the representational simplicity and gradient simplicity will give rise to the PRH.

\paragraph{Main result.} What we aim to prove in this note is the following theorem.
\begin{theorem}\label{theo: main}
    (Perfect Platonic Representation Hypothesis \cite{ziyin2025neural}) We train $f_A$ on $\mathcal{D}_{Z_A}$ and $f_B$ on $\mathcal{D}_{Z_B}$. Let the width of $A$ and $B$ be no smaller than the rank of $V^*$. Let both networks be at any global minimum of Eq.~\eqref{eq: entropic loss}. Then, for any invertible $M^O_A, M^I_A, Z_A, M^O_B, M^I_B, Z_B$, for any possible pair of $i$ and $j$, 
    \begin{enumerate}[noitemsep,topsep=0pt, parsep=0pt,partopsep=0pt, leftmargin=25pt]
        \item $h^i_A$ and $h^j_A$ are {perfectly aligned};
        \item $h^i_B$ and $h^j_B$ are {perfectly aligned};
        \item $h^i_A$ and $h^j_B$ are {perfectly aligned}.
    \end{enumerate}
\end{theorem}
Part (3) is what one would usually call a PRH, and essentially implies parts (1) and (2). To summarize the setting and its meaning, this model is different from the standard deep linear network with the three additional arbitrary matrices $M^O,\ M^I,\ Z$ such that
\begin{itemize}[noitemsep,topsep=0pt, parsep=0pt,partopsep=0pt, leftmargin=25pt]
    \item the deep linear network $W_D\cdots W_1$ models a block of network module under consideration;
    \item the matrices $M^I$ and $M^O$ model the part of the architecture that comes after and before this specific module, and are different for different networks;\footnote{An important remaining question is to what extent these matrices can be made nonlinear. In some sense, the matrix $M^I$ can be regarded as a linearization of a nonlinear module around some cluster mean $x'$, and so the theory states that to first order in the variance of $x$, the model learns a universal representation. The matrix $M^O$ is more easily generalizable to nonlinear cases because a lot of empirical works have shown that the last layers of a well-trained neural network essentially behave like linear layers \cite{masarczyk2024tunnel}, and universal representations tend to emerge in the later layers.}
    \item the matrix $Z$ models different views/modalities of the same underlying data.
\end{itemize}
Taking these interpretations, the theorem thus states that any EDLN module in the same or two different networks will learn the same representation (different up to a rotation and scaling), even if the architectures of the models are different and even if they are trained on different forms of input data. Also, note that the two EDLNs do not have to have the same width or depth, and do not have to have the same initializations.

\section{Most global minima are not platonic}\label{sec: most solutions are not platonic}
It should first be emphasized that Theorem~\ref{theo: main} is rather extraordinary. Even for the simple model of EDLN, most of the global minima of $L$ are not Platonic.
Consider the empirical loss:
\begin{equation}
L(\theta)= \E ||M^OW_D\cdots W_1M^I x-y||^2,
\end{equation}
There are a lot of symmetries (transformations of $\theta$ that leave the loss function value unchanged) in the loss function. For example, if $(W_D^*,..., W_{i+1}^*, W_i^*,...,W_1^*)$ is a global minimum, then, 
\begin{equation}
    (W_D^*,..., W_{i+1}^* T, T^{-1}W_i^*,...,W_1^*)
\end{equation}
is also a global minimum for any invertible matrix $T$. Therefore, if $\theta^*_A = (W_D^*,..., W_{i+1}^*, W_i^*,...,W_1^*)$ has a representation that is perfectly aligned to the representations of $f_B$, one has
\begin{equation}
    h_A^\top(x_1^A)h_A(x_2^A) =  h_B^\top(x_1^B)h_B(x_2^B).
\end{equation}
After the transformation by $T$ on net $A$, the l.h.s. becomes 
\begin{equation}
    h_A^\top(x_1)(T^{-1})^{\top} T^{-1} h_A(x_2),
\end{equation}
whereas the r.h.s. remains unchanged. For a general $T$, these two layers can no longer be perfectly aligned. Therefore, there are infinitely many minimizers of $L$ that are not Platonic (in fact, almost all solutions are non-Platonic). This means that it is highly nontrivial for SGD training to only learn the Platonic solution, even for the simple EDLN model. 

\section{Proof}
We present the proof of Theorem~\ref{theo: main} in a rather pedagogical manner. Also, because the derivation is the same for net $A$ and net $B$, we focus on net $A$. The most important equations in the derivation are placed in a \boxed{box}.

First of all, it is obvious (or, a good exercise to show) that any global minimum of $\E_{x\sim \mathcal{D}_{Z_A}} [\ell]$ satisfies $W_D...W_1 = (M^O)^{-1} V^* (M^I)^{-1} Z_A^{-1}$, and this is achievable as long as the model's width is larger than ${\rm rank}(V^*)$. So, it suffices to identify the solutions $\theta$ that minimize
\begin{equation}
    \E[\|\nabla \ell\|^2] 
\end{equation}
subject to the constraint 
\begin{equation}\label{eq: L minimum}
    W_D...W_1 = (M^O)^{-1} V^* (M^I)^{-1} Z_A^{-1}.
\end{equation}

Thus, let us consider a solution that satisfies $W_D...W_1 = (M^O)^{-1} V^* (M^I)^{-1} Z_A^{-1}$. Now, we perform the following transformation of this solution such that it does not change $L$ but reduces $\E[\|\nabla\ell\|^2]$. Fix a layer $i$, and let 
\begin{equation}
    W_i \to e^{\lambda T} W_i, 
\end{equation}
\begin{equation}
    W_{i+1} \to  W_{i+1} e^{-\lambda T},
\end{equation}
where $T$ is an arbitrary \textit{symmetric} matrix. By standard results of Lie groups, $e^{-\lambda T}$ is full-rank, and its inverse is $e^{\lambda T}$. Thus, $L$ does not change after this transformation. How does the entropy term change?  

Let $W_i^\lambda = e^{\lambda T}W_i$, and $W_{i+1}^\lambda = W_{i+1}e^{-\lambda T}$. Then, by parameter symmetry, for any $\lambda$,
\begin{equation}
    \ell(W_i^\lambda , W_{i+1}^\lambda) = \ell(W_i, W_{i+1}).
\end{equation}
Taking derivative with respect to $W_i$ and $W_{i+1}$ and applying chain rule, we obtain:
\begin{equation}
\boxed{
    e^{\lambda T} \nabla_{W_i^\lambda} \ell(W_i^\lambda , W_{i+1}^\lambda) = \nabla_{W_i}\ell(W_i, W_{i+1}).}
\end{equation}
\begin{equation}
    \boxed{\nabla_{W_{i+1}^\lambda} \ell(W_i^\lambda , W_{i+1}^\lambda) e^{-\lambda T}  = \nabla_{W_{i+1}}\ell(W_i, W_{i+1}).}
\end{equation}
Thus, the gradient is not invariant to the symmetry transformation. We will see in the following derivation that it is this equivariance of the gradient with respect to the symmetry transformation that leads to the Platonic representation. 


Applying these equations, one can show that this transformation only affects the part of $S$ that is due to $\nabla_{W_i}\ell$ and $\nabla_{W_{i+1}}\ell$. Thus,
\begin{equation}
    S = \Tr[\E[\nabla_{W_i}\ell \nabla_{W_i}^\top\ell]]  +  \Tr[\nabla_{W_{i+1}}\ell \nabla_{W_{i+1}}^\top\ell]  + \text{ other terms.}
\end{equation}
This transformation changes the two terms to 
\begin{equation}\label{eq: two terms}
    \Tr[ e^{-2\lambda T}\E\left[\nabla_{W_i}\ell \nabla_{W_i}^\top\ell] \right]  +  \Tr[ e^{2\lambda T} \E\left[ \nabla_{W_{i+1}}\ell \nabla_{W_{i+1}}^\top\ell]\right].
\end{equation}
Observe (!!!) that this equation reaches its minimum at a unique $\lambda = \lambda^*$ (which can be checked by taking the derivative of $\lambda$ and that its derivative is a monotonic function that passes through zero or extrapolates towards zero). This means that there exists a unique $\lambda$ for which the entropy is minimized. This is a general result for any Lie-group symmetries (e.g., see Theorem 4.3 of Ref.~\cite{ziyin2024parameter} and Theorem 4 of Ref.~\cite{ziyin2025neural}). Formally, the $\lambda$ that minimizes $S$ can be found by taking the derivative of $S$ with respect to $\lambda$ and set to zero:
\begin{align}
    \frac{d}{d\lambda} S &\propto \Tr[ - T e^{-2\lambda T}\E\left[\nabla_{W_i}\ell \nabla_{W_i}^\top\ell] \right]  +   \Tr[ T e^{2\lambda T} \E\left[ \nabla_{W_{i+1}}\ell \nabla_{W_{i+1}}^\top\ell]\right] \\
    &=  \Tr[ - T \E\left[\nabla_{W_i^\lambda}\ell \nabla_{W_i^\lambda}^\top\ell] \right]  +   \Tr[ T\E\left[ \nabla_{W_{i+1}^\lambda}\ell \nabla_{W_{i+1}^\lambda}^\top\ell]\right]\\
    &=0.
\end{align}
Equivalently, at any global minimum (or local minimum), the following equation must be satisfied for any symmetric $T$:
\begin{equation}\label{eq: balance condition}
\boxed{
    \Tr[T \E\left[\nabla_{W_i^\lambda}\ell \nabla_{W_i^\lambda}^\top\ell] \right]  =   \Tr[ T\E\left[ \nabla_{W_{i+1}^\lambda}\ell \nabla_{W_{i+1}^\lambda}^\top\ell]\right]}.
\end{equation}
Now, one can consider two types of $T$. The first type is a diagonal $T$ such that for a fixed index $j$,
\begin{equation}
    T_{kl} = \begin{cases}
    1 & \text{if $k=l=j$;}\\
    0 & \text{otherwise.}
    \end{cases}
\end{equation}
Note that this transformation is identical to the standard ``rescaling symmetry." Using its definition, one immediately arrives at a ``gradient" balance condition:\footnote{Note the following interesting fact: for an EDLN or a ReLU network, gradient regularization leads to gradient balance, whereas weight decay leads to a norm balance, and these two balancing conditions must trade off with each other in reality.}
\begin{equation}\label{eq: diagonal gradients}
     {\E\left[ \|\nabla_{W_i^{j:}}\ell \|^2\right]} = {\E\left[ \|\nabla_{W_{i+1}^{:j}}\ell \|^2\right]}
\end{equation}
where $W_i^{j:}$ is the $j$-th row of $W_i$, and $W_{i+1}^{:j}$ is the $j$-th column of $W_{i+1}$, and this condition holds for every $i$ and $j$.\footnote{To see why this is the unique minimizer of $S$, note that for this symmetry, the two terms in Eq.~\eqref{eq: two terms} simplifies to:
\begin{equation}
     \Tr[ e^{-2\lambda T}\E\left[\nabla_{W_i}\ell \nabla_{W_i}^\top\ell] \right] = e^{-2\lambda} \E\left[ \|\nabla_{W_i^{j:}}\ell \|^2\right] + \text{ constants in $\lambda$},
\end{equation}
\begin{equation}
     \Tr[ e^{-2\lambda T}\E\left[\nabla_{W_{i+1}}\ell \nabla_{W_{i+1}}^\top\ell] \right] = e^{2\lambda} \E\left[ \|\nabla_{W_{i+1}^{:j}}\ell \|^2\right] \text{ constants in $\lambda$},
\end{equation}
where $W_i^{j:}$ is the $j$-th row of $W_i$, and $W_{i+1}^{:j}$ is the $j$-th column of $W_{i+1}$. The sum of these two terms is minimized uniquely at:
\begin{equation}
    e^{4\lambda}  =  \frac{\E\left[ \|\nabla_{W_i^{j:}}\ell \|^2\right]}{\E\left[ \|\nabla_{W_{i+1}^{:j}}\ell \|^2\right]}.
\end{equation}}

The second type of $T$ we choose is a symmetric matrix with exactly two off-diagonal nonvanishing terms. For two fixed but arbitrary indices $j\neq j'$,
\begin{equation}
        T_{jj'} = T_{jj'} = 1,
\end{equation}
and all the other elements are zero. Plug this definition into Eq.~\eqref{eq: balance condition}, we obtain
\begin{equation}
    {\E\left[ \nabla_{W_i^{j:}}^\top\ell \nabla_{W_i^{j':}}\ell \right]} = {\E\left[ \nabla_{W_{i+1}^{:j}}^\top\ell \nabla_{W_{i+1}^{:j'}}\ell \right]}.
\end{equation}
This result, together with Eq.~\eqref{eq: gradient balance}, can be written as a concise matrix form:
\begin{equation}
    \E\left[ \nabla_{W_i^{j:}}\ell \nabla_{W_i^{j':}}^\top\ell \right] = \E\left[ \nabla_{W_{i+1}^{:j}}\ell \nabla_{W_{i+1}^{:j'}}^\top\ell \right].
\end{equation}
Now, define $K = W_{i+1} W_i$, and using the chain rule, 
\begin{equation}
   W_i \nabla_{K}^\top \ell  = \nabla_{W_i^{j:}}\ell,
\end{equation}
we can obtain a suggestive equation that any (local or) global minimizer of Eq.~\eqref{eq: entropic loss} must satisfy:\footnote{Note that this equation is very general and independent of the actual forms of the loss function or the architecture. It can be proved to hold whenever the loss function has the matrix rescaling symmetry: $
\ell(W,U) = \ell(WU)$. This equation, for example, holds for the query and key weight matrices within any self-attention layer in a transformer. Therefore, this equation is really only a consequence of parameter symmetry and entropic forces. For a general theory of how the entropy term changes in the presence of Lie-group parameter symmetries, see Ref.~\cite{ziyin2025neural}.}
\begin{equation}\label{eq: gradient balance}
   \boxed{ W_i \E[ \nabla_{K}^\top \ell \nabla_{K} \ell  ] W_i^\top = W_{i+1} \E[ \nabla_{K} \ell \nabla_{K}^\top \ell  ] W_{i+1}^\top,}
\end{equation}
where $K = W_{i+1} W_i$. Simply using the chain rule leads to
\begin{equation}
    \nabla_{K}^\top \ell = \nabla_{h_{i+1}}\ell h_{i-1}^\top,
\end{equation}
and so 
\begin{equation}
    \E[\nabla_{K}^\top \ell \nabla_{K} \ell]  = \E[\|\nabla_{h_{i+1}}\ell\|^2 h_{i-1} h_{i-1}^\top].
\end{equation}
The quantity
\begin{equation}
    \nabla_{h_{i+1}}\ell =      W_{i+2}^\top...W_D^\top (M^O)^\top (f_A(x) - y)  =  W_{i+2}^\top...W_D^\top (M^O)^\top \epsilon,
\end{equation}
where we have used the relation $f_A(x) - y = \epsilon$ at the global minimum~\eqref{eq: L minimum}, and so this term is independent of $h_{i-1}$ because of the independence between $\epsilon$ and $x$. Thus,
\begin{equation}
    \E[\nabla_{K}^\top \ell \nabla_{K} \ell] = \E[\| \nabla_{h_{i+1}}\ell\|^2] \E[h_{i-1} h_{i-1}^\top].
\end{equation}
A similar argument leads to
\begin{equation}
    \E[\nabla_{K} \ell \nabla_{K}^\top \ell] = \E[\|h_{i-1}\|^2] \E[ \nabla_{h_{i+1}}\ell\nabla_{h_{i+1}}^\top\ell].
\end{equation}
Together, this means that Eq.~\eqref{eq: gradient balance} can be written as\footnote{Note that this equation can be written as $\E[h_i h_i^\top] \propto \E[\nabla_{h_i} \ell \nabla_{h_i}^\top \ell]$ and so this step also proves the representation-gradient alignment hypothesis in Ref.~\cite{ziyin2024formation} for the EDLN model.}
\begin{equation}\label{eq: gradient balance 2}
    a_h W_i \E[h_{i-1} h_{i-1}^\top] W_i^\top =  a_g W_{i+1}^\top \E[\nabla_{h_{i+1}} \ell \nabla_{h_{i+1}} \ell^\top] W_{i+1}.
\end{equation}
where $a_h = 1/\E[\|[h_{i-1}\|^2]$ and $a_g = 1/ \E[\| \nabla_{h_{i+1}}\ell\|^2] $.

We can define $\bar{M}^I:= W_{i-1} ... W_1 M^I Z_A$ and $\bar{M}^O:= M^O W_D W_{i+2} $ so that 
\begin{equation}
    h_{i-1} = \bar{M}^I  x,
\end{equation}
\begin{equation}
    \nabla_{h_{i+1}} \ell =  (\bar{M}^O)^\top \epsilon.
\end{equation}
Thus, Eq.~\eqref{eq: gradient balance} can be written as
\begin{equation}\label{eq: layer conditions}
    a_h W_i \bar{M}^I \Sigma_x (\bar{M}^I)^\top W_i^\top = a_g W_{i+1}^\top (\bar{M}^O)^{\top} \Sigma_\epsilon \bar{M}^O W_{i+1}.
\end{equation}

Together with Eq.~\eqref{eq: L minimum}, we have obtained two equations to solve for two unknown matrices $W_i$ and $W_{i+1}$.\footnote{Here, assuming that both matrices are square matrices with dimension $d^2$, there are $2d^2$ many degrees of freedom to be determined. The first equation has $d^2$ constraints, and the second equation has $d(d+1)/2$ constraints. Together, this implies that there are $d(d-1)/2$ remaining degrees of freedom, which (as it turns out) corresponds to the rotation degrees of freedom between $W_{i+1}$ and $W_i$. However, these degrees of freedom do not affect the universal alignment at all and so are benign for our purpose.} This set of equations can be simplified greatly if we define
\begin{equation}
    \bar{W}_i =  {W}_i \bar{M}^I \sqrt{\Sigma_x}
\end{equation}
\begin{equation}
    \bar{W}_{i+1} = \sqrt{\Sigma_\epsilon} (\bar{M}^O)^\top {W}_{i+1}.
\end{equation}
From which we obtain two very simple equations to solve:
\begin{equation}\label{eq: set of equations 2}
\boxed{
    \bar{W}_{i+1} \bar{W}_i   =  \sqrt{\Sigma_\epsilon} V^* \sqrt{\Sigma_x} := \bar{V},}
\end{equation}
\begin{equation}\label{eq: set of equations 2b}
    \boxed{a_h \bar{W}_i  \bar{W}_i^\top =  a_g \bar{W}_{i+1}^\top \bar{W}_{i+1}.}
\end{equation}
The solution to this set of equations can be written using the SVD of $\bar{V}=E_l \Sigma E_r$, where $E_l$ and $E_r$ are the singular vectors and $\Sigma$ is a positive diagonal matrix of all the singular values. The solutions that satisfy this set of equations can be exhaustively written as
\begin{equation}
    \bar{W}_i =   R \sqrt{S} E_r,
\end{equation}
\begin{equation}
    \bar{W}_i =   E_l \sqrt{S} R^\top,
\end{equation}
where $R$ is an arbitrary orthogonal matrix, which does not affect the universality of the solution.

For our purpose, it is sufficient to note that the solutions of $\bar{W}_i$ and $\bar{W}_{i+1}$ are completely independent of $M^I, M^O$ and $Z_A$, and once we determine $\bar{W}_i$, we can determine $W_i$.

Now, what is the representation $h_i$?
\begin{equation}
    h_i(Z_A x) =   W_i \bar{M}^I x = \bar{W}_i \sqrt{\Sigma_x}^{+}x,
\end{equation}
where the $^+$ superscript denotes the pseudoinverse. 
But recall that $\bar{W}_i$ is completely independent of $M^I, M^O$ and $Z_A$ (up to a scalar constant $a_h/a_g$), and so $h_i$ must be \textit{independent of these matrices} (!!!). This equation is also independent of the layer index $i$, and so we have proved the first result: every layer of $f_A$ must be aligned with every possible layer of $f_A$.

The same argument applies to net $B$, whose layers all obey the same equation in Eq.~\eqref{eq: set of equations 2} and \eqref{eq: set of equations 2b} (which are \textit{independent of the network index $A$ and $B$}). This proves the second and third parts of the theorem: every layer of $A$ is aligned to every layer of $B$. Thus, we have proved Theorem~\ref{theo: main}.

\begin{remark}
    A key tool in the proof is the Lie-group symmetries in a deep linear network. While the proof we presented here does not rely on the learning dynamics, it may be possible to do an alternative proof by establishing the convergence of SGD to Platonic solutions. One possibility is to leverage the stochastic differential equation formalism of Ref.~\cite{ziyin2024parameter} to prove the convergence.
\end{remark}

It is also worthwhile to comment on this derivation from the perspective of conventional numerical analysis. In numerical analysis, it is well known that the numerical integrators tend to solve an alternative ``modified" problem that is perturbatively away from the original. In our theory, the entropic loss in Eq.~\eqref{eq: entropic loss} is exactly the modified problem being solved by the SGD algorithm (and $\eta$ is the perturbative parameter), and Eq.~\eqref{eq: entropic loss} can be seen as the modified energy functional of the continuous gradient flow when integrated with Euler discretization. It is an established wisdom in numerical analysis that it is these backward errors (the entropy term) that dominate the long-term behavior of the integrator, especially when there are symmetries in the dynamical variables \cite{hairer2006geometric}. Thus, it is no surprise that these integrators select very special solutions from a large degenerate manifold of solutions.

The network we studied is a deep linear model, which raises the question of whether linearity is necessary for PRH. There are two possible ways to interpret this result: (1) perfect PRH is achievable even in nonlinear models, and there is a hidden mechanism that is shared between deep linear and nonlinear models that leads to this effect (one such mechanism could be overparametrization); (2) perfect PRH is only achievable in deep linear networks, and so the closeness to perfect alignment can be seen as a metric of how close the representation is to a linearized representation. This might be a particularly relevant hypothesis because linearized structures are found to emerge quite often in later layers of neural networks. That being said, whether this is the case or not is not answerable in our current framework, and is an interesting open problem. 

\section{What breaks the perfect plato?}

This theory suggests an interesting new perspective for understanding and studying the PRH. Conventionally, one assumes that having no alignment is the default and asks why and when one gets a positive alignment. However, through the perfect PRH, one can think of perfect alignment as the default scenario, and ask the question of what breaks one away from this perfection, which could be much easier to study.\footnote{Just like how deviations from the ideal gas are much easier to study than deviations from a, say, strongly interacting fluid with a complicated equation of state, deviations from a simple ``idealized” learning system are far more tractable than a fully realistic, messy system.} In this section, we take this new perspective and tentatively answer the question of ``when is the perfect PRH broken?"

\paragraph{Weight Decay Breaks PRH.} One thing that breaks the perfect alignment between two layers is the use of weight decay. This can be understood easily (if we are hand-wavy). For simplicity, consider training two models while setting both of their $M^O$ and $M^I$ to be identity, and we have two different $Z_A\neq Z_B$ applied to transform the input data $x$. If we train with a small but positive weight decay and with gradient flow (so that the entropic term $\eta S$ is exactly zero),  we are really minimizing the following objective:
\begin{equation}\label{eq: wd loss}
    \min_{\theta: L(\theta)=\Tr[\Sigma_\epsilon]} {\|\theta\|^2}.
\end{equation}
For illustration and for conciseness, let us also focus on the case where $Z_A$ and $Z_B$ commute with $V^*$ and with each other, and let us also assume that they are all positive semidefinite matrices so that we can write things like $Z_A^{1/3}$, but the conclusion still holds without these assumptions. They are really there to make the equations easy to read. Now, the solution of Eq.~\eqref{eq: wd loss} for the EDLN model is (and up to rotations that do not matter)
\begin{equation}
    W_D =... = W_1 = (V^* Z_A^{-1})^{1/D}.
\end{equation}
Essentially, this is because the use of weight decay causes different layers to have a balanced norm. Now, consider the first-layer representation, for example,
\begin{equation}
    h^i = W_i ... W_1  Z_A x =  (V^* Z_A^{-1})^{i/D}Z_A x = (V^*)^{i/D} Z_A^{\frac{D-i}{D}}x.
\end{equation}
Therefore, as long as $i\neq D$, the representation always depends on $Z_A,$ which is an arbitrary transformation. Thus, the latent representations of network $A$ depend on $Z_A$, and those of $B$ depend on $Z_B$ for all the latent layers. This directly implies that these layers cannot be perfectly aligned. A more technical proof of this can be found in Ref.~\cite{ziyin2025neural}.

\paragraph{Gradient Flow Breaks PRH.} Now, assume that we are really training with the gradient flow (GF) algorithm on $L$ (instead of training in the zero learning rate limit of GD or SGD):
\begin{equation}
    \dot{\theta} = - \nabla_\theta L (\theta).
\end{equation}
This training procedure is thus not regularized by gradient or by weight decay. Now, because of the ``double rotation" symmetries of the EDLN loss function, a large set of conserved quantities\footnote{Namely, functions of $\theta$, $g(\theta)$ that do not change during training: $\frac{d}{dt} f(\theta)=0$.} exist: for all $i$,
\begin{equation}
    \frac{d}{dt}(W_{i+1}^\top W_{i+1} -  W_{i} W_{i}^\top) = 0,
\end{equation}
which implies that 
\begin{equation}
    W_{i+1}^\top (t) W_{i+1} (t) = W_{i} (t) W_{i}^\top(t) + A_0,
\end{equation}
where $A_0$ is the difference between these two matrices at initialization and can be arbitrary. This means that the spectra of $W_i$ and $W_{i+1}$ are dependent on their initializations. This result is rather well-known and has been used in a lot of prior works to analyze the learning dynamics of neural networks \cite{du2018algorithmic}.

The existence of these conserved quantities essentially implies that GF converges to solutions that are strongly initialization-dependent. Thus, if networks $A$ and $B$ have different initializations, they will not share perfectly aligned representations. In contrast, when the training proceeds with SGD, these quantities are no longer conserved, and \textit{it is the breaking of these conservation laws due to entropy that leads to the PRH}.

\paragraph{Label Transformations Breaks PRH.} As the theorem implies, transformations of the input data do not break the PRH, and so one naturally wonders if the learned representations are also invariant to full-rank transformations of the label $y$. It turns out that, surprisingly, transformations of the label indeed break the PRH. To see this, let us consider the simple case of a two-layer linear network. We assume that all six matrices $M^O$, $M^I$, $Z$ are identities, and the  network is trained on different views of the same label:
\begin{equation}
    \mathcal{D}_A = \{(x_i, \Phi_A y_i )\}_i,
\end{equation}
\begin{equation}
    \mathcal{D}_B = \{(x_i, \Phi_B y_i )\}_i,
\end{equation}
where $\Phi_A$ and $\Phi_B$ are invertible symmetric matrices. With this data, the global minimum for net $A$ transforms to :
\begin{equation}
    V^* \to  \Phi_A,
\end{equation}
and the noise spectrum becomes to
\begin{equation}
    \Sigma_\epsilon' =  \Phi_A \Sigma_\epsilon' \Phi_A.
\end{equation}
We can apply Eq.~\eqref{eq: layer conditions} to the case $D=2$ to obtain that at any global minimum,
\begin{equation}
    W_2^\top \bar{\Sigma'}_\epsilon W_2 = W_1 \bar{\Sigma}_x W_1^\top,
\end{equation}
where $\bar{\Sigma}_\epsilon = {\Sigma}_\epsilon/\Tr[{\Sigma}_\epsilon]$ and $\bar{\Sigma}_x = {\Sigma}_x/\Tr[{\Sigma}_x]$ are the normalized second moment matrices of data. Defining new parameters $\bar{W}_1 = {W}_1 \sqrt{\Sigma_x}$ and $\bar{W}_2 =  \sqrt{\Sigma_\epsilon}\Phi_A {W}_2$. We obtain two equations to solve:
\begin{equation}
    \bar{W}_2^\top \bar{W}_2 = \bar{W}_1 \bar{W}_1^\top,
\end{equation}
\begin{equation}
    \bar{W}_2 \bar{W}_1 = \sqrt{\Sigma_\epsilon} \Phi_A^2 V^*.
\end{equation}
The solution is not independent of the transformation $\Phi_A$. In sharp contrast, the set of equations \eqref{eq: set of equations 2} and \eqref{eq: set of equations 2b} we had in the proof is transformation-independent. This implies that $\Phi_A$ directly influences the weight representation $W_1x$ and that the representation is no longer Platonic after the $\Phi_A$ transform. This is an interesting theoretical prediction that has not yet been empirically confirmed.

\paragraph{Convergence to Saddle or Local Minima Breaks the PRH.} Apparently, if an EDLN converges to a saddle instead of a global minimum, it cannot be perfectly aligned to another EDLN that converges to the global minimum. For EDLN, the saddle points are the low-rank approximations $V^*$ \cite{fukumizu2000local}. One can show that these saddle points all have non-negative alignment to each other and to the global minimum, but it is no longer the case that the alignment will be perfect. 

\paragraph{Data Discrepancy / Heterogeneity Breaks the PRH.} For example, suppose $x_i^A = x_i + z_i^A$ for some additional and unique feature $z_i^A$ while $x_i^B = x_i + z_i^B$ for some additional and unique feature $z_i^B$. Then, it is obvious that at the global minimum, as long as the labels do not depend only on $x_i$, the learned representation will not be universal. This agrees with the findings in Ref.~\cite{tjandrasuwita2025understanding}, for example.

\paragraph{Edge of Stability (EOS) breaks PRH.} It turns out that the same derivation we have shown here can be used to prove the progressive sharpening phenomenon \cite{cohen2021gradient}, and, surprisingly, within the EDLN model, the force (due to entropy) that gives rise to the progressive sharpening is exactly the force that gives rise to the Platonic representations. As the models learn universal representations, they tend to move towards a gradually sharper loss landscape, eventually reaching a fixed point (namely, the unique minimizers of the Lie-group symmetries discussed below Eq.~\eqref{eq: two terms}). This is because the solution a model learns is independent of data (in the case of PRH), whereas the sharpness of a solution is dependent on the data (say, on the $Z_A$ matrix). Therefore, for a very ill-conditioned $Z_A$, the Platonic solution can be arbitrarily sharp -- thus, converging towards the Platonic solution leads to progressive sharpening.\footnote{More precisely, depending on the data distribution, the same force can also lead to progressive flattening. Therefore, an empirical prediction of the theory is that PRH always happens together with progressive sharpening and/or progressive flattening. Also, it is worthwhile to clarify the phenomenology of EOS. The EOS really is a combination of two independent phenomena: (1) progressive sharpening, where a neural network moves to sharper and sharper places during training, and (2) proper EOS, where the sharpness stays at a critical value and stops increasing further. Our theory shows that progressive sharpening is due to entropy, while the proper EOS is not due to leading-order entropic effects.}

However, SGD training can only take the model so far as the ``edge of stability" because higher-order effects due to discretized time steps tend to suppress sharpness. When the Platonic solution is very sharp, SGD training cannot reach it due to the need to stay at the edge of stability. There is a lot more depth to this part, which is no longer directly related to the proof of the PRH, and we refer the interested readers to Ref.~\cite{ziyin2025neural} for solving the EDLN model for progressive sharpening and its intriguing relationship with the PRH.

\section{Discussion}\label{sec: discussion}

In this note, we have presented an exactly solvable model that exhibits the perfect PRH. This model suggests that it is the discretization error of training and gradient noise that determines the representation learning, consistent with the observation made in Ref.~\cite{ziyin2024formation}. We now discuss various implications of the theory in more detail.

\vspace{-1mm}
\paragraph{Cause of the Platonic Representation Hypothesis.} From a pure machine learning perspective, Ref.~\cite{huh2024platonic} conjectured three possible mechanisms that give rise to the PRH: (a) increasing capacity, (b) simplicity bias of training, or (c) multitasking training. Interestingly, the mechanism studied here does not really match these previously conjectured mechanisms. Our mechanism has nothing to do with multitasking.  Moreover, our result holds for any (embedded) deep linear network, all having the same capacity and the same level of simplicity, because all solutions parametrize the same input-output map. Here, the cause of the universal representation is the parameter symmetry alone: in the degenerate manifold of solutions, the training algorithm prefers a particular and universal one. Our example highlights how symmetry has been an overlooked fundamental mechanism in deep learning. 

\vspace{-1mm}
\paragraph{Imperfect Platonic Representation Hypothesis.} In this work, we proved the most strict and idealized version of the PRH, where \textit{all possible layers become mutually and perfectly aligned}. Experimentally, this has indeed been confirmed to happen for deep linear networks in Ref.~\cite{ziyin2025neural}. We have also studied how the solution will break away from this perfect PRH. The perfect PRH is a very strong condition. The fact that it can be proved makes it possible to prove weaker forms of it. In reality, the PRH only holds imperfectly, where the data are weakly but significantly aligned. For example, if measured with CKA, the perfect PRH is satisfied when the CKA alignment score is exactly one, whereas, for real models, this value has been measured to be somewhere around $0.2$ \cite{huh2024platonic}. Thus, extending our results to more realistic settings of nonlinear models and weaker forms of PRH is a crucial future theoretical step.


\vspace{-1mm}
\paragraph{Emergent Invariances in Learning Algorithms.} Our theory implies that optimization-unrelated (i.e., due to the implicit regularization effect of the entropic force) learning dynamics may give rise to models that share universal features with each other. What we have focused on is SGD, and a remaining question is to what extent this is true for other learning algorithms. Also, we note that the learned representation is not fully universal in the sense that the learning representation is invariant to variations of the input data (modeled by the $Z$ matrices) but is not invariant to the label, as discussed in the previous section. This raises the interesting open question of whether it is possible to have algorithms that learn representations that are also invariant to the transformations of the label. Such a global minimum certainly exists for deep linear networks. For example, if we let the first layer be $W_1= \Sigma_x^{-1}$, then the representation of the first layer will be both invariant to the input data from and to the label; the question is whether it is possible to find learning algorithms that achieve this. This also implies a hierarchy of universality: at the very top, there are learning algorithms whose solutions are almost invariant to every variation of the data (maybe in addition to other things), and below it, there are algorithms whose learning representations are either invariant to input or output, and at the bottom of the hierarchy are those algorithms that are invariant to neither (such as gradient flow).

\vspace{-1mm}
\paragraph{Universal Phenomena and Unified Theories.} A rather striking point we made in the previous section is that two seemingly unrelated universal phenomena in deep learning, (1) PRH and (2) progressive sharpening, can have the same hidden cause and can be explained by a unified theory of entropic forces. This raises the hope of being able to find unified explanations of more interesting and intriguing phenomena in deep learning. In particular, this raises the interesting question of whether or not other seemingly unrelated phenomena in deep learning can be found to share a hidden root. Identifying these connections will be a big step forward towards building the theory of deep learning.

\vspace{-1mm}
\paragraph{Feature Learning and Infinitesimal Learning Rate.} In the context of the NTK and feature learning, a common result states that for an infinitesimal learning rate (and under certain scaling), SGD converges to the solution of kernel regression. Let $\theta$ denote the solution found by SGD at time $t$ and with learning rate $\eta$; this result states that
\begin{equation}
    \lim_{t \to \infty} \lim_{\eta \to 0^+}\theta(t,\eta) = \theta^*_{\text{kernal regression}}.
\end{equation}
This argument is often made assuming that the order of taking these two limits does not matter. However, a key aspect of the proof of our theorem is that these two limits are not equal:
\begin{equation}
    \theta^*_{\text{kernal regression}} = \lim_{t \to \infty} \lim_{\eta \to 0^+}\theta(t,\eta) \neq \lim_{\eta \to 0^+} \lim_{t \to \infty} \theta(t,\eta) =\  ?
\end{equation}
and cannot be exchanged. The l.h.s. completely ignores the entropic effect, whereas the r.h.s. takes that into account. This could hint at a fundamental problem with understanding AI training with continuous time formalisms, such as in the NTK \cite{jacot2018neural} and feature learning \cite{yang2020feature} limit. Arguably, the second limit is much closer to reality because of, for example, the ubiquity of the edge of stability phenomenon, which cannot be approximated by the simple gradient flow. 

\vspace{-1mm}
\paragraph{More Technical Implications.} While our derivation ended with the solution of Eq.~\eqref{eq: set of equations 2} and \eqref{eq: set of equations 2b}, these solutions can be solved in much more depth to give an explicit solution of the global minimum of the EDLN model. This solution can be shown to have many other interesting properties. For example, one can show that all the intermediate layers must become rotation matrices and have the same norm as each other. This explains why all the intermediate layers are aligned with each other even within the same model -- it is because the intermediate layers do not do much meaningful computation. At the same time, the first and last layers will be special and do most of the interesting computations. In particular, the first layer will convert the representation to a universal form such that it does not depend on the covariance of $x$, in effect whitening the data. The first and last layers will also have different norms and spectrums from all the intermediate layers. This could imply that there are interesting dynamics that only happen at the ``surface" of neural networks, resembling the surface modes and edge modes common in physical systems. Thus, one might be interested in developing a theory of surface physics for neural networks. We refer the readers to Refs.~\cite{ziyin2024parameter} and \cite{ziyin2025neural} for more discussion of these technical properties.

\section*{Acknowledgement} 

The authors thank Yizhou Xu, Tomaso Poggio, Brian Cheung, and Phillip Isola for discussion during the writing of this note. ILC acknowledges support in part from the Institute for Artificial Intelligence and Fundamental Interactions (IAIFI) through NSF Grant No. PHY-2019786. 

\bibliographystyle{plain}
\bibliography{ref}

\end{document}